\documentclass{article}
\PassOptionsToPackage{numbers, compress}{natbib}

\usepackage{xr-hyper}

\usepackage[preprint]{neurips_data_2021}

\usepackage[utf8]{inputenc} 
\usepackage[T1]{fontenc}    
\usepackage{hyperref}       
\usepackage{url}            
\usepackage{booktabs}       
\usepackage{amsfonts}       
\usepackage{nicefrac}       
\usepackage{microtype}      
\usepackage[dvipsnames]{xcolor}         
\usepackage{siunitx}        

\usepackage{todonotes}
\usepackage{subcaption}

\usepackage{amsmath}
\usepackage{mathtools}  
\usepackage{amsfonts}

\DeclareMathOperator*{\argmax}{arg\,max}        
\newcommand{\MDP}[0]{\mathcal{M}}                       
\newcommand{\statespace}[0]{\mathcal{S}}                
\newcommand{\actionspace}[0]{\mathcal{A}}               
\newcommand{\transdomain}[0]{\mathcal{T}}               
\newcommand{\rewards}[0]{\mathcal{R}}                   

\newcommand{\policy}[0]{\pi}                            

\newcommand{\inst}[0]{i}                    
\newcommand{\insts}[0]{\mathcal{I}}         

\newcommand{\cMDP}[0]{\MDP_\insts}                      
\newcommand{\cMDPdef}[0]{\cMDP \coloneqq \{\MDPi\}_{\inst \sim \insts}} 
\newcommand{\MDPi}[0]{\MDP_\inst}                       
\newcommand{\MDPidef}[0]{\MDPi \coloneqq (\statespace, \actionspace, \transdomaini, \rewardsi)}  

\newcommand{\transdomaini}[0]{\transdomain_\inst}       
\newcommand{\rewardsi}[0]{\rewards_\inst}               

\usepackage{wrapfig}
\usepackage{xspace}
\usepackage{pifont}
\usepackage{xcolor}

\newcommand{\tick}{\textcolor{ForestGreen}{\ding{52}}}
\newcommand{\ok}{\textcolor{Dandelion}{\ding{108}}}
\newcommand{\cross}{\textcolor{BrickRed}{\ding{56}}}

\newcommand{\GenRL}{CARL\xspace} 

\makeatletter
\newcommand*{\addFileDependency}[1]{
  \typeout{(#1)}
  \@addtofilelist{#1}
  \IfFileExists{#1}{}{\typeout{No file #1.}}
}
\makeatother

\newcommand*{\myexternaldocument}[1]{%
    \externaldocument{#1}%
    \addFileDependency{#1.tex}%
    \addFileDependency{#1.aux}%
}
\myexternaldocument{standalone_appendix}
\myexternaldocument{standalone_appendix}

\title{\GenRL: A Benchmark for\\ Contextual and Adaptive Reinforcement Learning} 
\author{%
Carolin Benjamins\thanks{Equal Contribution, Contact Author}$^{\;\:1}$, Theresa Eimer$^{\tiny{*}1}$, Frederik Schubert$^{1}$, André Biedenkapp$^{2}$,%
\and%
\textbf{Bodo Rosenhahn$^{1}$, Frank Hutter$^{2\:3}$} and \textbf{Marius Lindauer$^{1}$}\\
$^1$Leibniz University Hanover $^2$University  of  Freiburg $^3$Bosch Center for Artificial Intelligence\\
\texttt{$\left\{\right.$benjamins,eimer$\left.\right\}$@tnt.uni-hannover.de} \
}

\begin{document}

\maketitle

\begin{abstract}
While Reinforcement Learning has made great strides towards solving ever more complicated tasks, many algorithms are still brittle to even slight changes in their environment. 
This is a limiting factor for real-world applications of RL.
Although the research community continuously aims at improving both robustness and generalization of RL algorithms, unfortunately it still lacks an open-source set of well-defined benchmark problems based on a consistent theoretical framework, which allows comparing different approaches in a fair, reliable and reproducible way.
To fill this gap, we propose \GenRL, a collection of well-known RL environments extended to contextual RL problems to study generalization.
We show the urgent need of such benchmarks by demonstrating that even simple toy environments become challenging for commonly used approaches if different contextual instances of this task have to be considered. Furthermore, \GenRL allows us to provide first evidence that disentangling representation learning of the states from the policy learning with the context facilitates better generalization.
By providing variations of diverse benchmarks from classic control, physical simulations, games and a real-world application of RNA design, \GenRL will allow the community to derive many more such insights on a solid empirical foundation.
\end{abstract}
\section{Introduction}\label{sec:intro}




Reinforcement Learning (RL) has driven progress in areas like game playing \citep{silver-nature16a,badia-icml20}, robot manipulation \citep{lee-sciro20}, traffic control~\citep{arel-its10a}, chemistry~\citep{zhou-acs17a} and logistics~\cite{li-aamas19a}. 
At the same time, RL has shown little to no success in real-world deployment in important areas such as healthcare or autonomous driving. 
This is largely explained by the fact that modern RL agents are often not primarily designed for generalization, making them brittle when faced with even slight variations in their environment \citep{yu-corl19,lu-amia20,meng-data19}. 
Since we cannot assume that RL agents will be able to observe all kinds of states and transitions for varying instances of tasks, 
these agents need to become more adaptable and robust.

To address this limitation, there is increased interest in Meta-RL approaches, aiming to improve learning across different tasks~\citep{finn-icml17a,schulman-iclr16,duan-corr16,wang-cogsci17a,matiisen-ieee20,klink-neurips20,nguyen-ki21,eimer-icml21a}. 
The focus mostly lies on increasing the sample efficiency of agents, few-shot transfer of policies to new tasks or on solving harder tasks.
Similarly, Robust-RL addresses generalization to smaller variations in the environment by ensuring a stable performance under task modelling errors or noisy observations \citep{morimoto-neurips16,pinto-icml17,zhang-iclr21b}.
While these directions are important in making RL more broadly and robustly applicable, with \GenRL we aim for providing the foundations of more general RL agents. Optimally, these agents should be capable of zero-shot transfer to prior unseen environments and changes in transition dynamics while interacting with an environment~\cite{zhang-iclr21a,fu-icml21a,yarats-icml21a,abdolshah-icml21a,sodhani-icml21a}.

Unfortunately, there is a lack of established benchmarks for studying the notion of generalization.
In fact, we often observed that researchers employed hand-crafted modifications to commonly accepted tasks to enable benchmarking of Meta-RL.
For example, multiple different modifications to the well known \emph{CartPole} task, in which an agent needs to learn to balance a pole on top of a movable cart, have been used in different publications to show generalization abilities of agents~\citep{seo-neurips20,kaddour-neurips20,eimer-icml21a}.
In particular, different pole lengths are used to study whether a general agent can balance poles that it has not seen during training.
However, the pole lengths or distributions over pole lengths vary in different publications, hindering comparisons and reproducibility.
With \GenRL we provide such distributions to facilitate better comparability and reproducibility for further research.


Our goal is to address all these issues by proposing \GenRL, a benchmark library allowing to reliably and reproducibly study general RL agents. To this end, \GenRL has well defined distributions and bounds over the space of environments to generalize to and poses a low barrier of entry in terms of compute.
To build on a sound theoretical foundation, we make use of the contextual Reinforcement Learning paradigm (cRL) \citep{hallak-corr15} and build contextual extensions to environments from the literature including OpenAIs Gym~\citep{gym} and the Brax physics engine \cite{brax2021github}.
The notion of context in the environment enables us to define a variety of tasks and distributions of tasks which an agent can encounter during training.
Changes in tasks can be as simple as defining different goal states, more complex by changing the transition dynamics (as the changes in pole length for the CartPole environment mentioned above) or a combination thereof, leading to varying levels of difficulty.
Most importantly and wherever possible, we base the notion of context on real-world physical properties, such as gravity, friction or mass of an object, see Figure~\ref{fig:fetch_cMDP} for an example of a contextually extended environment. Those properties are intuitive to understand and individually adjustable.

The proposed benchmark enables research on generalization capabilities of RL agents in cases where agents are explicitly or implicitly exposed to the context at hand but also in cases where the context is hidden and potentially has to be learned. In particular, we demonstrate the usefulness of our proposed \GenRL benchmark library by evaluating and discussing:
{
\begin{enumerate}
    \item The influence of varying context and the importance of contexts in deep RL by increasing learning efficiency via available knowledge of task variations,
    \item The generalization capability of trained agents to in-distribution environments,
    \item The generalization capability of trained agents to out-of-distribution environments, and
    \item The big next challenges for general RL which can be studied in principled way with \GenRL.
\end{enumerate}
}

\section{\GenRL's Theoretical Foundation: Contextual RL (cRL)}

\begin{figure}
    \centering
	\begin{subfigure}[t]{0.60\textwidth}
		\centering
        \includegraphics[width=0.8\textwidth]{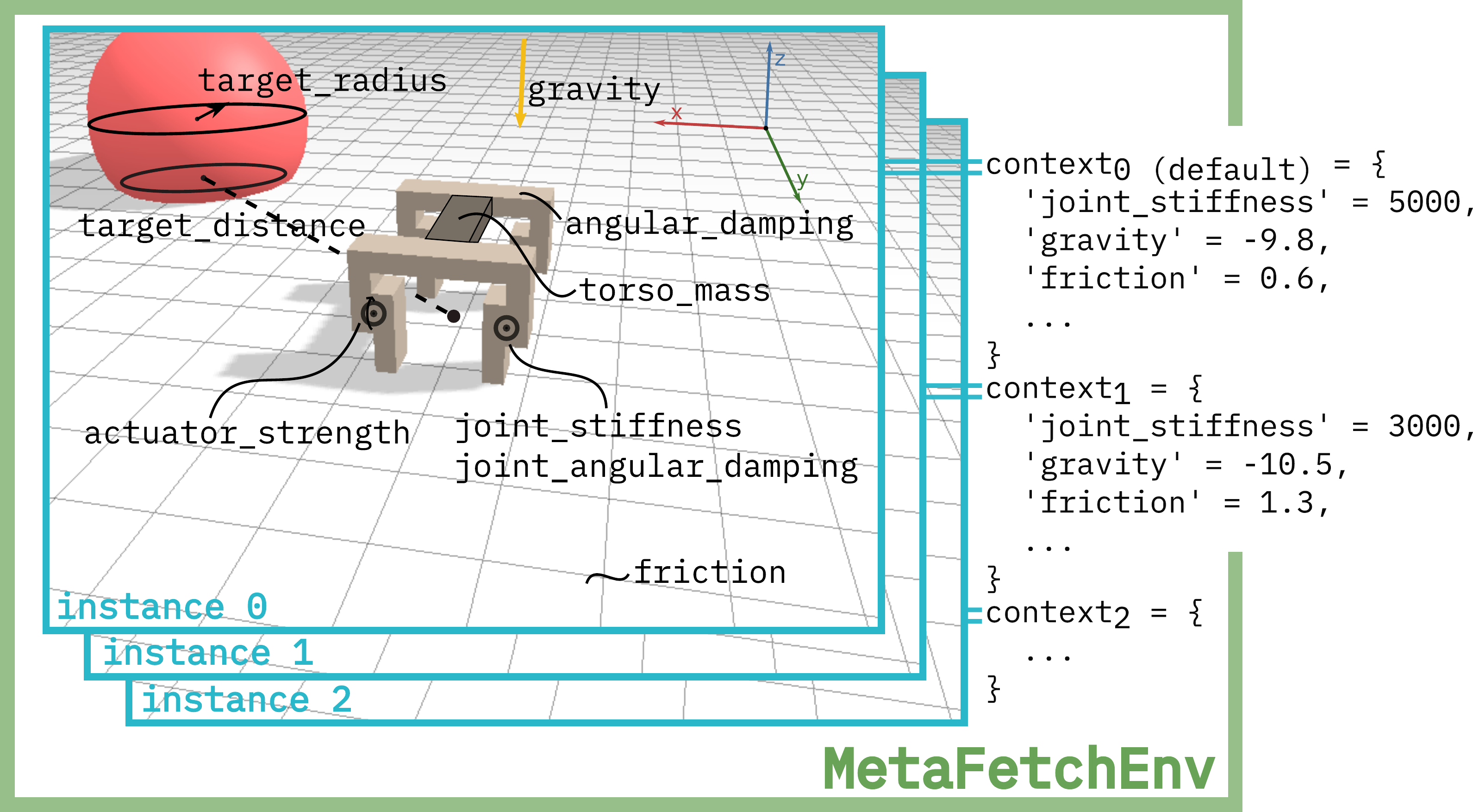}
        \caption{Example of a configurable \GenRL environment} \label{fig:fetch_cMDP}
	\end{subfigure}
	\quad
	\begin{subfigure}[t]{0.26\textwidth}
		\centering
    	\includegraphics[width=1\linewidth]{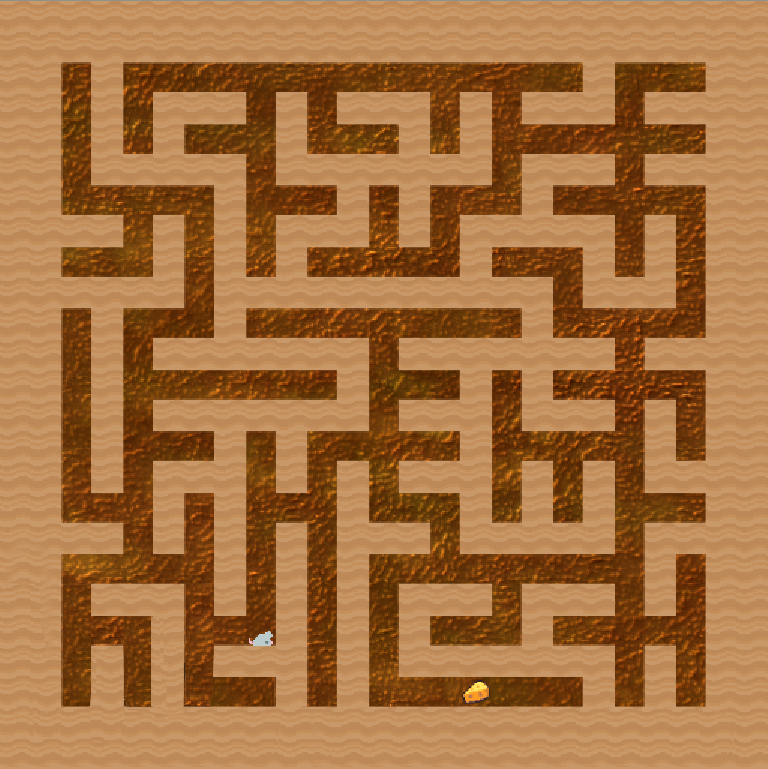}
        \caption{ProcGen Maze instance}\label{fig:procgenmazes}%
	\end{subfigure}
    \caption{(\protect\subref{fig:fetch_cMDP}) \GenRL makes the context defining the behavior of the environment visible and configurable. This way we can train for generalization over different instances (contexts) of the same environment. Here, we show all context features for Brax' Fetch~\cite{brax2021github} and sketch possible instantiations by setting the context features to different values. Fetch is embedded in the \GenRL environment controlling the instances. (\protect\subref{fig:procgenmazes}) A sampled ProcGen Maze instance.}

\end{figure}

A basic MDP is a $4$-tuple $\mathcal{M} \coloneqq (\statespace, \actionspace, \transdomain, \rewards)$ consisting of a state space $S$, an action space $A$, a transition function $\transdomain: \statespace \times \actionspace \to \statespace$ and a reward function $\rewards: \statespace \times \actionspace \to \mathbb{R}$. 
This abstraction, however, views reward and transition function as fixed, constraining the environment to a single instantiation without room for variations we would potentially see in real-world applications.
In the following we will refer to such a particular instantiation of an environment as an \emph{instance}.

\GenRL's theoretical foundation is build upon contextual RL. 
It extends the MDP formulation of RL problems to allow for the notion of several instances.
For example, an instance $\inst\sim\insts$ sampled from a distribution of instances $\insts$ could determine a different goal position in a maze problem (e.g., Figure~\ref{fig:procgenmazes}) or different gravity conditions (e.g., moon vs.~earth) for an airborne navigation task.
We refer to the information defining the instance at hand as the instance's context $c_\inst$. We note that $c_\inst$ is sometimes known to the agent (e.g., a broken leg), sometimes measured with noise (e.g., friction of floor), or maybe even completely unobservable (e.g., mass of an external object).
With \GenRL we provide a variety of such contexts that can influence an agent's learning and generalization capabilities.
We further provide bounds and distributions of these contexts to facilitate better reproducibility and comparability for future research.

In order to incorporate context into the MDP definition, we use contextual MDPs (cMDPs) \citep{hallak-corr15,modi-alt18,biedenkapp-ecai20}.
A contextual MDP $\cMDP$ is a collection of MDPs $\cMDPdef$ with $\MDPidef$. 
This formulation assumes a common state and action space as the underlying task stays the same; however, in each $\mathcal{M}_{\inst}$ an agent will potentially only be able to reach only a subset of states $S_{\inst} \subseteq S$. 
Transition and reward functions may vary between instances.

There are several different tasks of interest concerning cMDPs, all of which define an optimal policy~$\policy^*$ for a given cMDP in different ways. 
An example would be the focus on generalization performance where $\policy^*$ maximizes the expected return across a test set $\insts_{\text{Test}}$ drawn from the same distribution as the training set $\insts_{\text{Train}}$ with discount factor $\gamma$~\cite{biedenkapp-ecai20}:
\begin{equation}
    \policy^* \in \argmax_{\policy \in \Pi} \frac{1}{|\insts_{\text{Test}}|} \sum_{\inst \in \insts_{\text{Test}}} \sum_t^T \gamma^t R_i(s_t, \policy(s_t)),
\end{equation}
where $T$ corresponds to the maximal episode length.
In policy transfer, the focus is on the performance across a set of transfer instances specifically, which is often relatively small but rather heterogeneous.
Here, the test instance set $\insts_{\text{Test}}$ can largely differ from  $\insts_{\text{Train}}$, but the optimal policy would still maximize the mean reward across it just as above.
On the other hand, only the final performance on a single, very hard instance $\inst_H$ might be important and all other instances are only used to work towards that goal. That could be the case in, e.g., curriculum learning. In that case, we use the available training set $\insts_{\text{Train}}$ to find a $\policy^*$ such that:
\begin{equation}
    \policy^* \in \argmax_{\policy \in \Pi} \int_{\insts_{\text{Train}}} P(i) \sum_t^T \gamma^t R_{i}(s_t, \policy(s_t)) d i,
\end{equation}

where the probability $P(i)$ of considering an instance $i$ is skewed  towards $i_H$ over time.

cRL in \GenRL subsumes several other related formulations. For example, Goal-based RL \citep{florensa-icml18} uses the same idea of conditioning the reward function of each task on its specific goal, but is more limited in scope as the environment dynamics stay static throughout.
Block MDPs~\citep{du-icml19}, on the other hand, focus on state representations for generalization.
The task here is to learn a representation of the observable space of a family of environments that enables generalization across that family.
Just as in cRL, reward functions and transition dynamics both may vary with the family, but the focus is shifted away from learning a policy towards learning a meaningful representation.
While the original block MDP did not include specifics about how reward and transition functions differ within the environment family, contextual block MDPs provide the context as additional information~\citep{zhang-iclr21a}. 
As we have direct access to the context information on all \GenRL benchmarks, the base case provides context as in a cMDP.
However, users are free to switch to a hidden context version that requires a representation learning as in a block MDP. 

\section{Related Work}\label{sec:related_work}

\newcolumntype{P}[1]{>{\raggedright\arraybackslash}p{#1}}
\newcommand{\addlinespacetable}[0]{\addlinespace[0.01cm]}
\begin{table}[tbp]
    \small{
    \centering
    \caption{Comparison of the \GenRL Benchmark collection to related Benchmarks. Benchmarks are rated as true (\tick), somewhat true (\ok) or false (\cross) in each category.}
    \label{tab:comparison_benchmarks_related_work}
    \begin{tabular}{l cc cc c}
    \toprule
        Benchmark &  Open Source & Explicit Context & Cheap Training\footnotemark & Diverse Tasks & Varying $\mathcal{T} \& \mathcal{R}$ \\
        \midrule
        \addlinespacetable
        MDPP~\citep{rajan-arxiv19} 
        & \ok
        & \ok
        & \tick
        & \tick
        & \tick
        \\
        bsuite ~\citep{osband-iclr20} 
        & \tick
        & \cross
        & \tick
        & \tick
        & \cross
        \\
        \addlinespacetable
        ALE \citep{machado-jair28} 
        & \tick
        & \cross
        & \ok
        & \ok
        & \cross
        \\
        \addlinespacetable
        ProcGen \citep{cobbe-2019} 
        & \tick
        & \cross
        & \cross
        & \tick
        & \cross
        \\
        \addlinespacetable
        Alchemy \citep{wang-corr21} 
        & \tick
        & \ok
        & \cross
        & \ok
        & \tick
        \\
        \addlinespacetable
        Meta-world \citep{yu-corl19} 
        & \cross
        & \tick
        & \cross
        & \ok
        & \ok
        \\
        \addlinespacetable
        MTEnv \citep{sodhani-2021} 
        & \cross
        & \tick
        & \cross
        & \ok
        & \ok
        \\
        \addlinespacetable
        Safety Gym \citep{ray-2019} 
        & \cross
        & \cross
        & \cross
        & \ok
        & \tick
        \\
        \addlinespacetable
        TMA \citep{romac-icml21} 
        & \tick
        & \ok
        & \cross
        & \ok
        & \tick
        \\
        \addlinespacetable
        MiniGrid \citep{gym_minigrid}
        & \tick
        & \cross
        & \tick
        & \tick
        & \ok
        \\
        \addlinespacetable
        NetHack \citep{kuttler-neurips20}
        & \tick
        & \ok
        & \cross
        & \tick
        & \cross
        \\
        \addlinespacetable
        MiniHack \citep{samvelyan2021minihack}
        & \tick
        & \ok
        & \tick
        & \tick
        & \tick
        \\
        \midrule
        \addlinespacetable
        \GenRL (ours)
        & \tick
        & \tick
        & \tick
        & \tick
        & \tick
        \\
    \bottomrule
    \end{tabular}
    }
\end{table}








Benchmarks for generalization exist in different sub-fields of RL, each with its own focus.
MDP Playground \citep{rajan-arxiv19} and bsuite \citep{osband-iclr20} both contain small scale benchmarks intended to test specific qualities in RL algorithms (e.g., resistance to noise), both for the purpose of development and comparison between different algorithms.
In contrast, the focus of \GenRL is less on assessing RL algorithms against each other on fixed MDPs but in terms of their generalization capabilities to variations of MDPs. 
Benchmarks such as in MDP Playground and bsuite provide valuable feedback for researchers in development before they tackle more complex and opaque problems like the ones we provide.

In game simulations, the Arcade Learning Environment (ALE) has made an effort to include some challenges geared towards policy transfer and generalization in their ``flavours'' \citep{machado-jair28}.
However, the bigger challenge in this field is ProcGen \citep{cobbe-2019}.
It contains several arcade-style games with  procedurally generated level structures.
In a similar way, Alchemy \citep{wang-corr21} also provides a procedurally generated benchmark.
Even though it only contains a single task, this task is very complex compared to the games in ProcGen on their own.
Both are challenging benchmarks that require generalization from state observations only.
We believe that this approach is less valuable in many applications other than in games, because most often additional information is available. 
Additionally, while it is possible to specify levels with certain attributes in Alchemy, these procedurally generated benchmarks provide a far less fine-grained control over their context than the diverse set of benchmarks in \GenRL where users can directly specify their instances and control the similarity of their sampled contexts.
\GenRL{}'s flexibility allows for a better characterization of agents' generalization capabilities as well as the possibility of adding custom curricula for each environment.


Multi-task learning requires some amount of generalization, although here the focus is on accelerating the acquisition of skills on completely new tasks. 
For example, Meta-world \cite{yu-corl19} focuses on skill transfer in a few-shot setting, providing standardized test sets of different sizes. 
Its tasks are based on MuJoCo, however, which requires a paid license for large scale experiments, is comparatively much more expensive to run than the Brax physics simulator (used as part of \GenRL) and thus limits the accessibility of the benchmark.
Meta-world is also integrated in MTEnv~\cite{sodhani-2021}. 
MTEnv provides a strong benchmark for multi-task learning as well as representation learning. \GenRL can accommodate multi-task learning as well, but the focus is on the multitude of context options available in each of our environments and therefore generalization across different transition dynamics. 

There also exist more, related benchmarks in specialized subfields of RL. 
Safety Gym \cite{ray-2019} is targeted towards developing and testing algorithms for risk-sensitive domains. 
Also, TeachMyAgent~\cite{romac-icml21} is a benchmark for teacher-student based curriculum learning.
Both are well suited to the needs of their communities, but also narrow in their scope. 
While \GenRL currently does not provide explicit contexts for either safety or curriculum learning, it could be extended to cover both domains and will be especially relevant for any curriculum learning algorithms not using the teacher-student framework.

Overall, \GenRL is the only benchmark library that is completely open-source, allows for fine-grained control of context on a diverse set of benchmarks and thus allows to study the next generation of general RL agents in a reliable and reproducible way.
We summarize this comparison in Table~\ref{tab:comparison_benchmarks_related_work}\footnotetext{\label{note2}``Cheap Training'' here refers the total runtime of one agent. This takes into account both to the computational cost of the environment itself as well as the number of training steps necessary to expect results.}.


\section{The Role of Context in Deep RL and \GenRL}\label{sec:context_matters}

One important distinction that needs to be made in contextual RL concerns the ease of identifiability of context information.
Here we broadly distinguish between explicit contexts, i.e.~directly available information provided by the environment, and implicit contexts, i.e.~abstract information hidden in the available state.
While explicit contexts can directly be used by agents to infer the underlying transition dynamics, implicit contexts potentially need to be disentangled from the provided state.
In particular, we argue that deep RL research commonly already makes use of the notion of contexts.
This context however is only present in an implicit form in the state, thereby entangling representation learning capabilities with generalization capabilities of a policy.

For example in the ProcGen \emph{maze} environment (see Figure~\ref{fig:procgenmazes}),
an agent can observe the whole maze and is tasked with guiding a mouse from the bottom left corner to the cheese.
The maze structure, texture of the walkable tiles and the location of the goal (i.e.~cheese) are randomly generated for each new instance.
Note that the wall texture never changes and that observations are only available as images.
Thus, a capable RL agent could learn to directly extract the location of the mouse and cheese as well as classify which tiles are (not) walkable.
This extracted information then allows the agent to perform contextual RL.
In particular, \citeauthor{eimer-icml21a}~\cite{eimer-icml21a} showed that providing an agent with the coordinates of the agent and goal states as well as a flattened vector representation of a maze allows agents to make use of this context information to transfer behaviours between similar mazes.

A similar argument can be made for more complex environments where a ``level'' might not be fully observable.
For example, in the game Super Mario Bros., see Figure~\ref{fig:envs_mariorna}, Mario needs to reach the goal on the right side of the screen while avoiding enemies.
If an agent is made aware of the enemy types appearing in the level, through the use of context, this information can be used downstream in the policy net to learn appropriate offensive or defensive behaviour.
Another direct context feature could be an indicator which special ability an agent can use, potentially leading to different behaviour when Mario picked up a power-up.

We argue that benchmarks using procedural content generation are more suitable for evaluating the representation learning capabilities of agents rather than their ability to generalize.
In fact, the authors of ProcGen~\cite{cobbe-2019} used it to determine that the IMPALA-CNN~\cite{espeholt-icml18} architecture is more capable than the Nature-CNN~\cite{mnih-nature13} architecture for their considered setup.
Here, we propose benchmarks that provide a ground truth for the changes in underlying transition dynamics to study generalization while also containing more complex environments that can be used to study representation learning.
Disentangling these two important tasks will enable researchers to target each more efficiently and ultimately facilitates the development of new RL algorithms targeted towards generalization.
We use \GenRL to demonstrate that an agent making use of context information during training can learn to solve instances quicker and generalizes better than those that have to infer this information themselves (see Section~\ref{sec:hidden_vs_visible_context}).
This gives additional evidence that disentangling learning of such contextual features from learning the behaviour policy can improve the generalization capabilities of RL agents (see Appendix~\ref{subsec:repl} for further discussion).

\section{The \GenRL Benchmarks}

\begin{figure}	
	\centering
	\begin{subfigure}[t]{0.17\textwidth}
		\centering
		\includegraphics[width=1\textwidth]{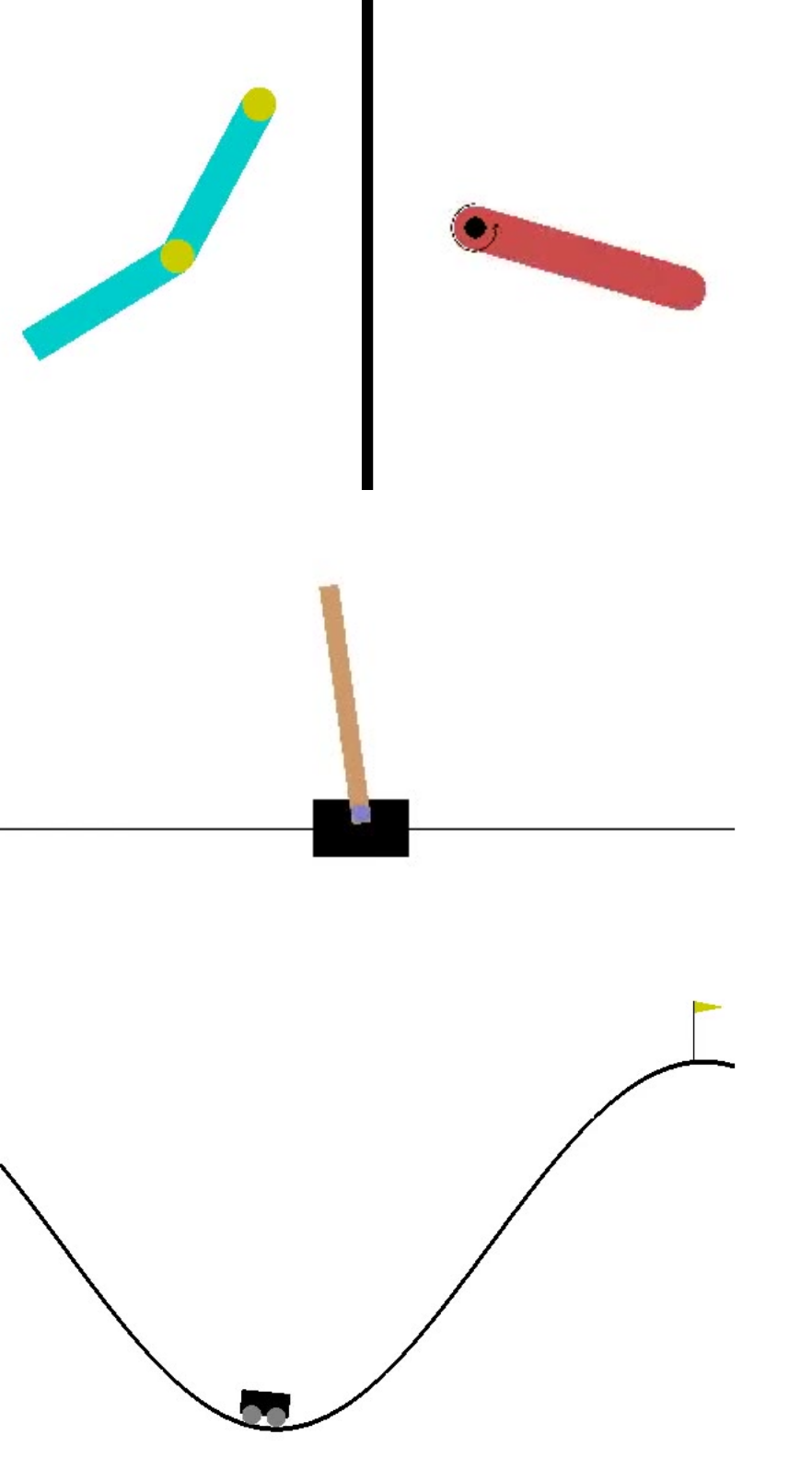}
		\caption{Classic Control~\cite{gym}}
		\label{fig:envs_classic_control}		
	\end{subfigure}
	\quad
	\begin{subfigure}[t]{0.15\textwidth}
		\centering
		\includegraphics[width=1\textwidth]{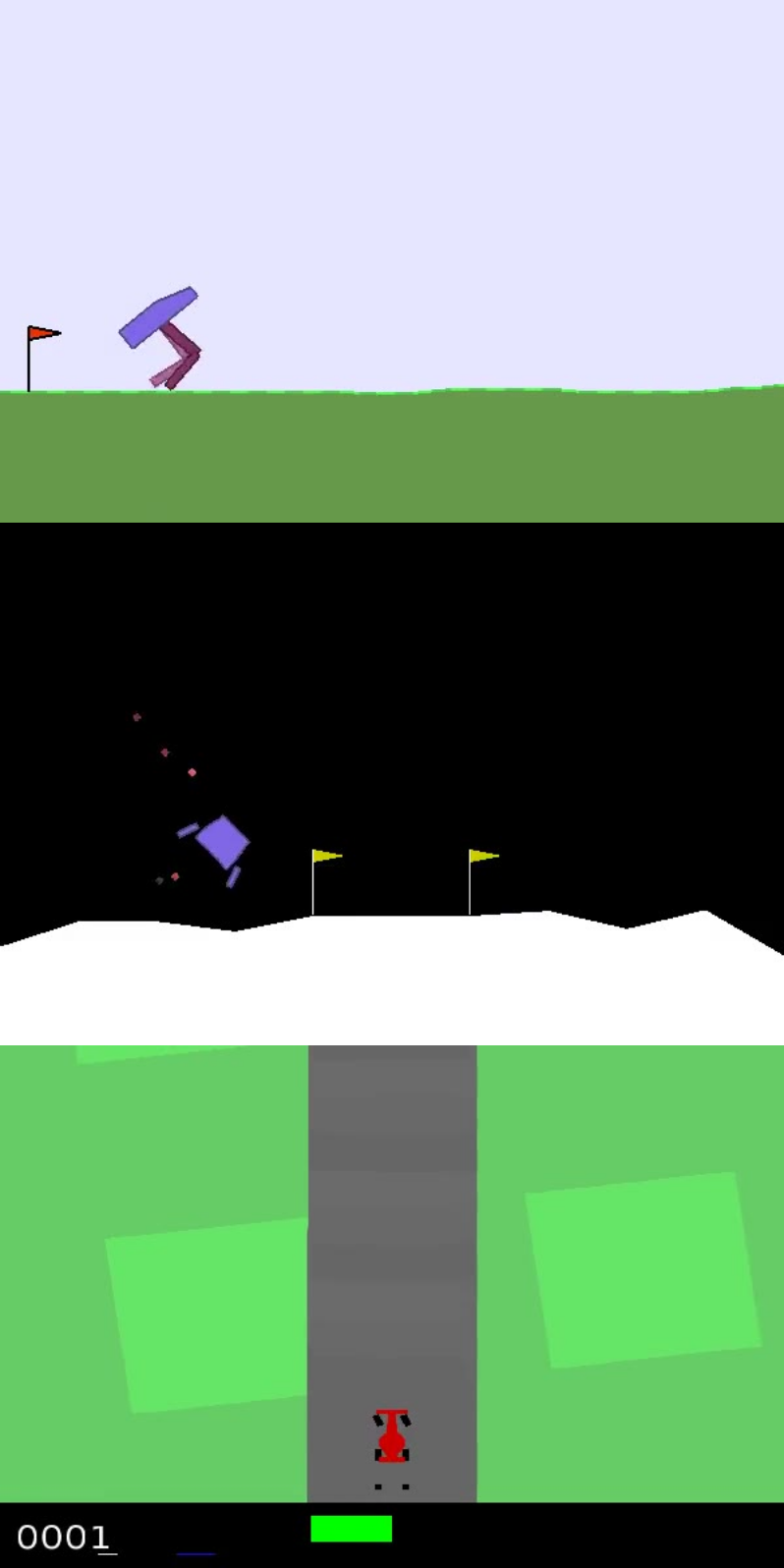}
		\caption{2D Physics Simulation~\cite{gym}}
		\label{fig:envs_box2d}		
	\end{subfigure}
	\quad
	\begin{subfigure}[t]{0.3\textwidth}
    	\centering
    	\includegraphics[width=1\textwidth]{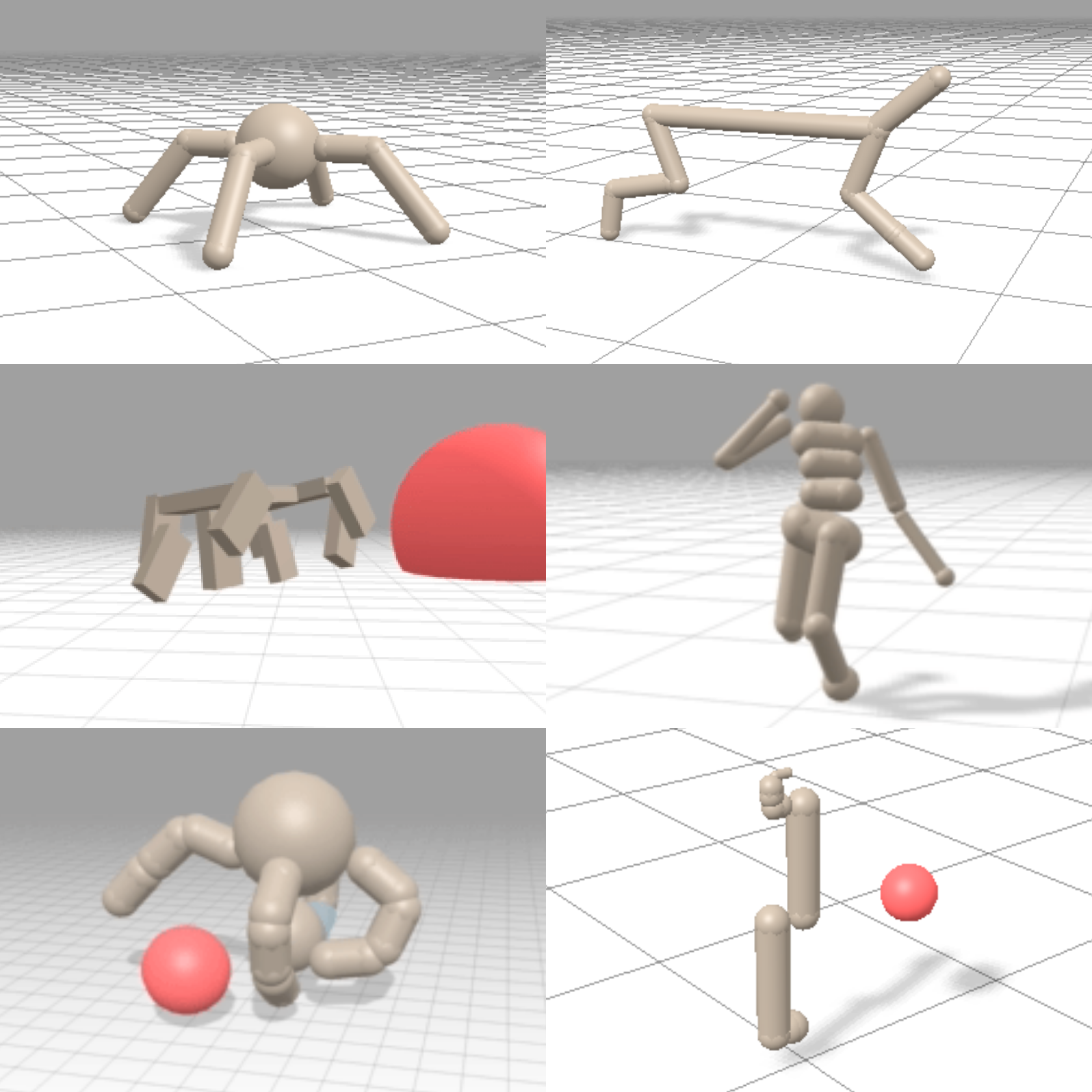}
    	\caption{Brax 3D Physics Simulation~\cite{brax2021github}}
    	\label{fig:envs_brax}		
	\end{subfigure}
	\quad
	\begin{subfigure}[t]{0.205\textwidth}
		\centering
		\includegraphics[width=1\textwidth]{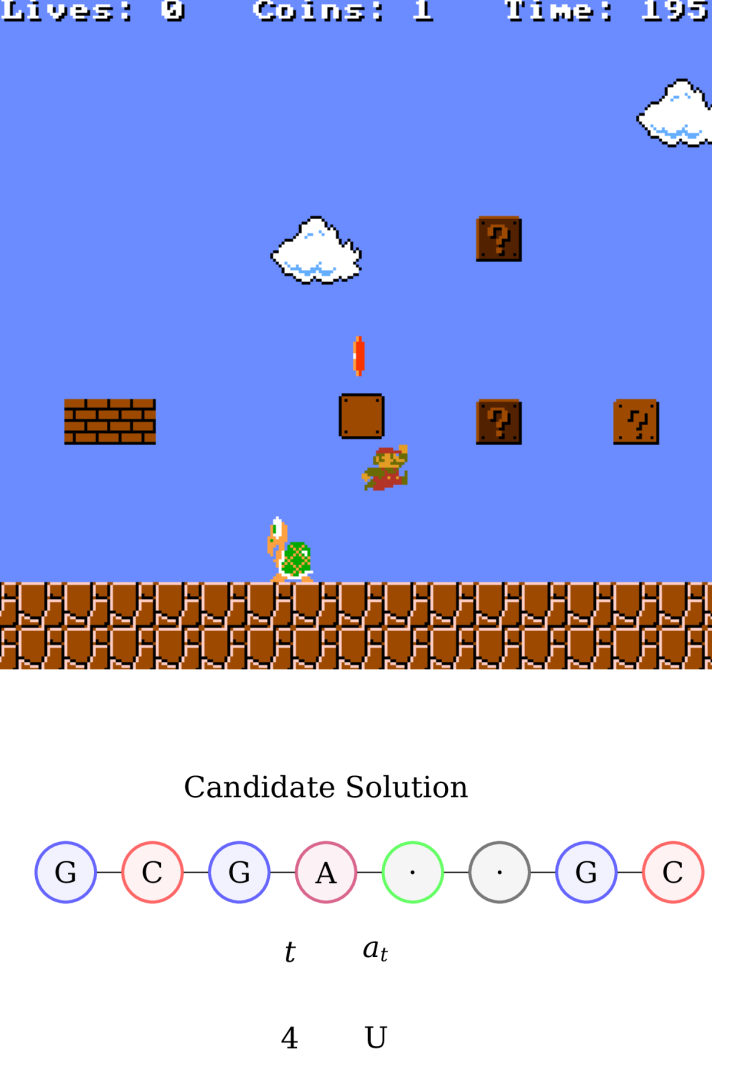}
		\caption{Super Mario~\cite{awiszus-aiide20,schubert-tg21} \& RNADesign~\cite{runge-iclr19a}}
		\label{fig:envs_mariorna}		
	\end{subfigure}
	\caption{\GenRL Environments; listed from top to bottom. (\protect\subref{fig:envs_classic_control}) OpenAI Gym's~\cite{gym} Acrobot and Pendulum, CartPole, MountainCar. (\protect\subref{fig:envs_box2d}) OpenAI Gym's~\cite{gym} BipedalWalker, LunarLander, CarRacing. (\protect\subref{fig:envs_brax}) Brax~\cite{brax2021github} Ant and HalfCheetah, Fetch and Humanoid, Grasp and UR5E. (\protect\subref{fig:envs_mariorna}) Super Mario~\cite{awiszus-aiide20,schubert-tg21} and RNADesign~\cite{runge-iclr19a}.}\label{fig:envs_overview}
\end{figure}

\begin{figure}[t]
    \centering
    \includegraphics[width=1\textwidth]{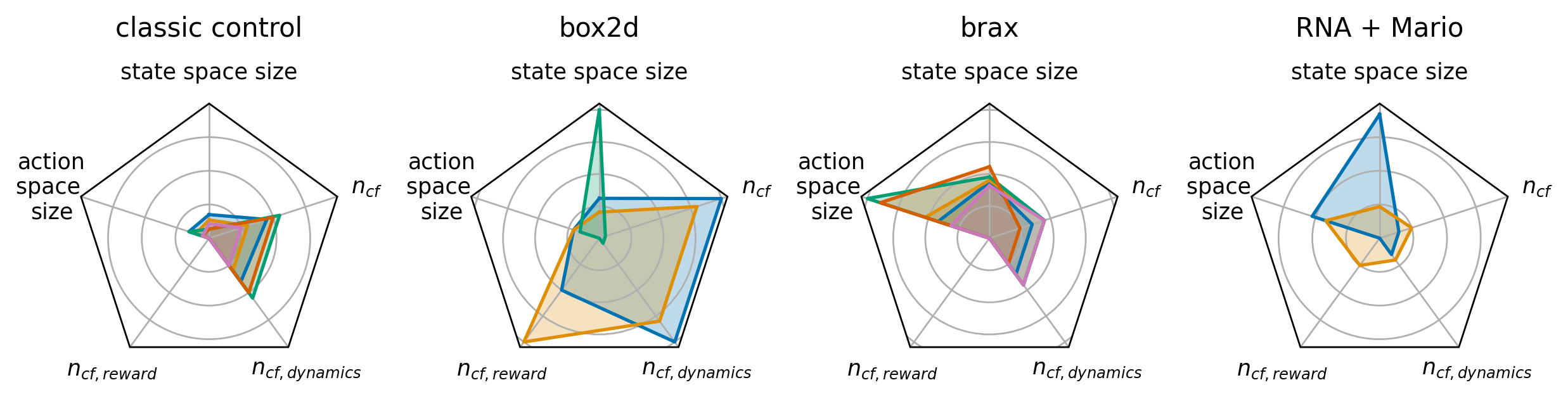}
    \caption{Characteristics of each environment of the environment families showing the action space size, state space size (log-scale), number of context features ($n_{cf}$), the number of context features directly shaping the reward ($n_{cf, reward}$) and the ones changing the dynamics ($n_{cf, dynamics}$). All axes are scaled to the global extrema and the state space size is additionally on a logarithmic scale.}
    \label{fig:radar_env_space}
\end{figure}

In order to gain insight on how the context and its augmentation influences the agent's learning and behavior, we provide several benchmarks in \GenRL.
As first benchmarks we include and contextually extend classic control and box2d environments from OpenAI Gym~\cite{gym}, Google Brax' walkers~\cite{brax2021github}, a RNA folding environment~\cite{runge-iclr19a} as well as Super Mario levels~\cite{awiszus-aiide20,schubert-tg21}. See Figure~\ref{fig:envs_overview} for an overview of included environments.
Although each environment has different tasks, goals and mechanics, the behavior of the dynamics and the rewards is influenced by physical properties.
A more detailed description of the environments is given in Appendix~\ref{appendix:cats}. In the following we will discuss the properties of the \GenRL Benchmarks which are summarized in Figure~\ref{fig:radar_env_space}.

\paragraph{State Space}
Most of our benchmarks have vector based state spaces that can either be extended to include the context information or not. 
The notable exceptions here are CARLVehicleRacing and CARLToadGAN, which exclusively use pixel-based observations.
The size of the vector based spaces range from only two state variables in the CARLMountainCar environments to 299 for the CARLHumanoid environment.

\paragraph{Action Space}
We provide both discrete and continuous environments, with six requiring discrete actions and the other ten continuous ones.
The actions range from a single dimension to 19.

\paragraph{Quality of Reward}
We cover different kinds of reward signals with our benchmarks, ranging from relatively sparse step penalty style rewards where the agent only receives a reward of $-1$ each step to complex composite reward functions in e.g.~the Brax-based environments.
The latter version is also quite informative, providing updates on factors like movement economy and progress towards the goal whereas the former does not let the agents distinguish between transitions without looking at the whole episode. 
Further examples for sparse rewards are the CARLCartPoleEnv and CARLVehicleRacingEnv.


\paragraph{Context Spaces}
While the full details of all possible context configurations can be seen in Appendix~\ref{sec:appendix_env_context_features}, for brevity here we only discuss the differences between context spaces and the configuration possibilities they provide.
Depending on the environment the context features have different influence on the dynamics and the reward. Of all $131$ registered context features, $\SI{98}{\percent}$ influence the dynamics. This means that if a context feature is changed the transition from one state into the other is changed as well. Only $\SI{5}{\percent}$ of the context features shape the reward. Most context features ($\SI{87}{\percent}$) are continuous, the rest is categorical or discrete.
With the explicit availability of context features \GenRL lends it self to study the robustness of agents by adding noise on top of the specific context features.
Further, the provided bounds and sampling distributions of the context spaces that are provided as part of \GenRL enable better comparability and reproducibility for future research efforts in the realm of general RL agents.

\paragraph{Summary}
Comparing our benchmarks along these attributes, we see a wide spread in most of them (Figure~\ref{fig:radar_env_space}). 
For the first iteration of \GenRL, we focused on fairly cheap-to-run problems to lower the barrier of entry as much as possible. Nevertheless, as \GenRL will further grow over time, the diversity of benchmarks will further increase and we will also include 
harder benchmarks.
Already now, \GenRL provides a benchmarking collection that tasks agents with generalizing in addition to solving the tasks most common in modern RL while providing a platform for reproducible research.




\section{Experiments}\label{sec:experiments}
Having discussed \GenRL's theoretical foundation as well as its initial set of benchmarks, 
we now study several first research questions regarding the effects of context.
Our experiments are designed to demonstrate that we can use \GenRL to gain meaningful insights into the Meta-RL setting even on simple environments. Details about the hyperparameter settings and used hardware for all experiments are listed in Appendix~\ref{appendix:hardhyper}.
In each of them, we train and evaluate on $5$ different random seeds and a set of $100$ sampled contexts.
All experiments can be reproduced using the scripts we provide with the benchmark library at \mbox{\url{https://www.github.com/automl/CARL}}.

\begin{figure}[ht]
    \centering
	\begin{subfigure}[t]{0.49\textwidth}
		\centering
        \includegraphics[width=1\textwidth]{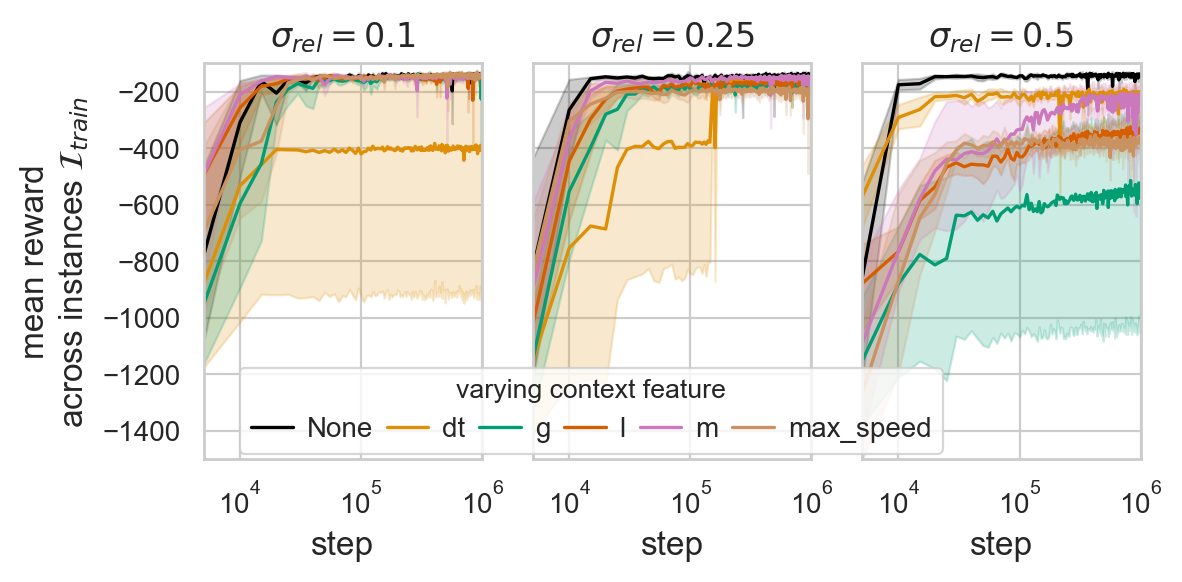}
        \caption{Q1: Different context distributions}
        \label{fig:results_CARLPendulumEnv_contexthidden}
 	\end{subfigure}
	\begin{subfigure}[t]{0.49\textwidth}
		\centering
    	\includegraphics[width=1\textwidth]{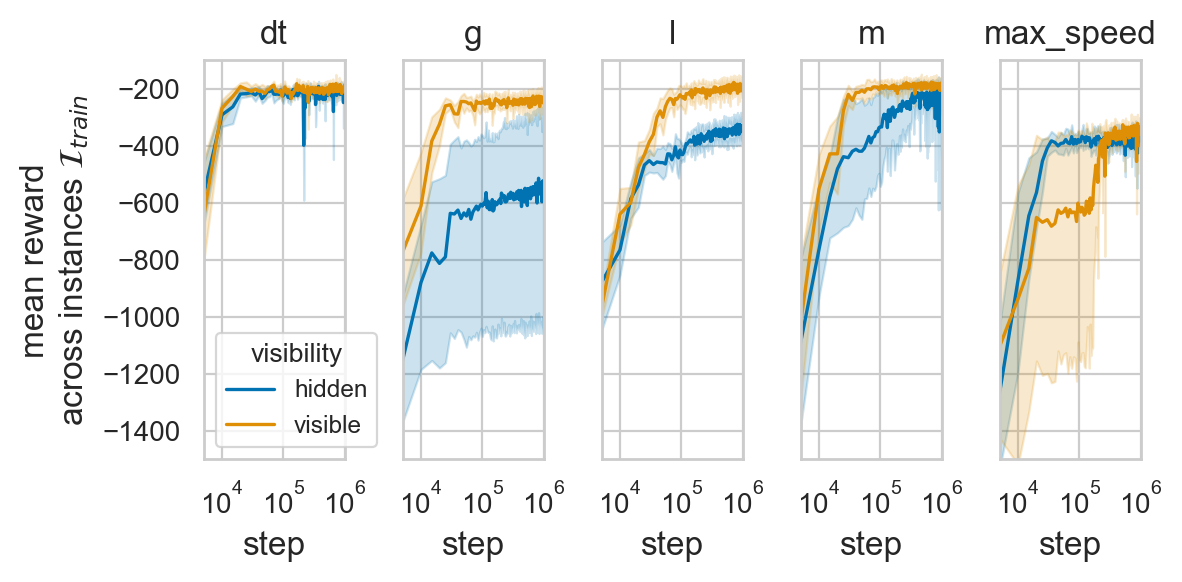}
        \caption{Q2: Visible vs. hidden context for $\sigma_\text{rel}=0.5$}\label{fig:results_CARLPendulumEnv_contextvisible}%
	\end{subfigure}
    \caption{Training performance of a DDPG agent on CARLPendulumEnv with (\protect\subref{fig:results_CARLPendulumEnv_contexthidden})
    different context distributions (Q1) and (\protect\subref{fig:results_CARLPendulumEnv_contextvisible}) the effect of visible context (Q2). $\sigma$ is the standard deviation for sampling the context. The context feature \texttt{dt} refers to the observation interval length, \texttt{g} to gravity, \texttt{l} to the pole length, \texttt{m} to the pole mass and \texttt{max\_speed} to the maximal speed of CARLPendulumEnv.}
    \label{fig:results_CARLPendulumEnv}
\end{figure}

\subsection{Q1: How do Context Features Influence an Agent's Training Performance?}
\label{sec:context_influence}
In order to gain an intuition on how context features influence an agent's training performance, we evaluate a DDPG~\cite{lillicrap-iclr16} agent on the well known \emph{Pendulum} task from OpenAI gym~\cite{gym}.
Through \GenRL, we can vary the context features \emph{gravity (g)}, \emph{integration time step (dt)}, pendulum \emph{mass (m)} and \emph{length (l)} as well as the \emph{maximal speed} (see Figure~\ref{fig:results_CARLPendulumEnv}). 
In Equation \ref{sec:pendulum_dyna_eq} in the appendix, we show the dynamic system of Pendulum.

To understand how context features influence an agent's performance, for each considered context feature we sample a set of $100$ instances $\{\inst_k\}_{ k=1, \dots, 100}$ for each task within the ranges provided by the environment specification while keeping the others context features fixed to their default.
Each context feature $c$ of an instance is sampled from a normal distribution, centered around its default value such that $
    c \sim \mathcal{N}(c_{\text{def}}, \sigma_{\text{rel}} \cdot c_{\text{def}})$
where $c_{\text{def}}$ is the default value defined in the original environment and $\sigma_{rel}$ is the relative standard deviation.
Here we evaluate three relative standard deviations $\sigma_{\text{rel}} \in \{0.1, 0.25, 0.5\}$ to show the impact of varying similarities of the instance distribution.

Further, we treat the context as hidden, only implicitly noticeable to the agent through the observation of the state features.
While small changes in the context barely have an impact on the training performance of the agent, large variations of a \emph{single context feature} can make the learning task challenging (see Figure~\ref{fig:results_CARLPendulumEnv_contexthidden}).
This gives evidence that even simple, cheap-to-run environments can provide an agent with challenging learning tasks, depending on the level of generalization required.
Note, this style of training with implicit contexts is currently the default setting for training on vision-based environments such as ProcGen (see Section~\ref{sec:context_matters}).
We refer to a the appendix Section~\ref{sec:additional_results} for a first impression on the influence of context on a vision-based environment, CARLMarioEnv, showing similar insights as for Pendulum.

\subsection{Q2: Are Explicit Context Features Necessary to Learn General Agents?}
\label{sec:hidden_vs_visible_context}
To answer this question we first use the same agent and environment setup, i.e. DDPG with the same hyperparameters on Pendulum and the widest context distribution ($\sigma_\text{rel}=0.5$).
%
Our results (see Figure~\ref{fig:results_CARLPendulumEnv_contextvisible}) suggest that explicitly making agents aware to the change in transition dynamics generally results in a better performance when a generalization over strong deviations in context features is required.
This is clearly observable by comparing results for context features that have a higher impact on the final performance, such as \emph{gravity (g)}, \emph{pole length (l)} and \emph{pole mass (m)}. 
For a fairly low impact context feature \textit{integration time step (dt)}, making the context visible results in a lower standard deviation and a slightly higher final reward. Still, varying \textit{dt} led to minor loss of reward compared to original Pendulum task (black curve in Figure~\ref{fig:results_CARLPendulumEnv_contexthidden}).
For the \textit{max\_speed} context, both training variants struggle to achieve as high a reward.
In the early training stages, the agent trained with access to the context achieved a lower reward than its counterpart. However, in the latter half it could catch up and slightly improve over the context-oblivious agent.

\begin{wrapfigure}{R}{0.6\textwidth} 
    \centering
    \includegraphics[width=0.6\textwidth]{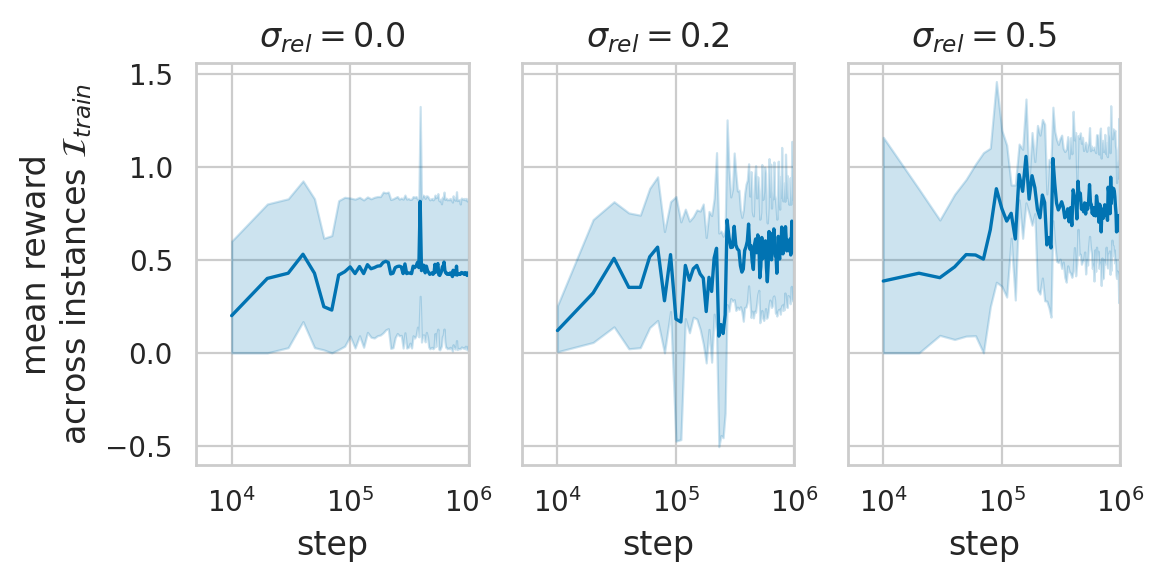}
    \caption{Training performance on CARLMarioEnv where the inertia of Mario is varied .}
    \label{fig:mario_inertia}
\end{wrapfigure}
One context feature that heavily influences the dynamics in CARLMarioEnv environment is the inertia of Mario.
In Figure~\ref{fig:mario_inertia}, we see that a higher variation of the inertia improves the performance of the PPO agent and leads to faster training.
This effect can be explained by the influence of Mario's inertia on the exploration of the agent (i.e. a lower inertia makes it easier for the agent to move).

An interesting question for future work is how different context features change the learning behavior of agents and to which degree generalization is impacted by it.

\subsection{Q3: To What Extent Can We Transfer a Learned Policy to a New Context?}
\label{sec:policy_transfer}
To gain insights to what extend the agent is able to transfer a learned policy to a new context we create the ``Landing in Space'' scenario based on the well known LunarLander environment.
To this end, a DQN~\cite{mnih-nature15} agent is trained on a rather narrow context distribution. For testing, we then place this agent in a new context which might not have been observed during training.

\begin{figure}[ht]
    \centering
	\begin{subfigure}[t]{0.49\textwidth}
		\centering
        \includegraphics[width=1\textwidth]{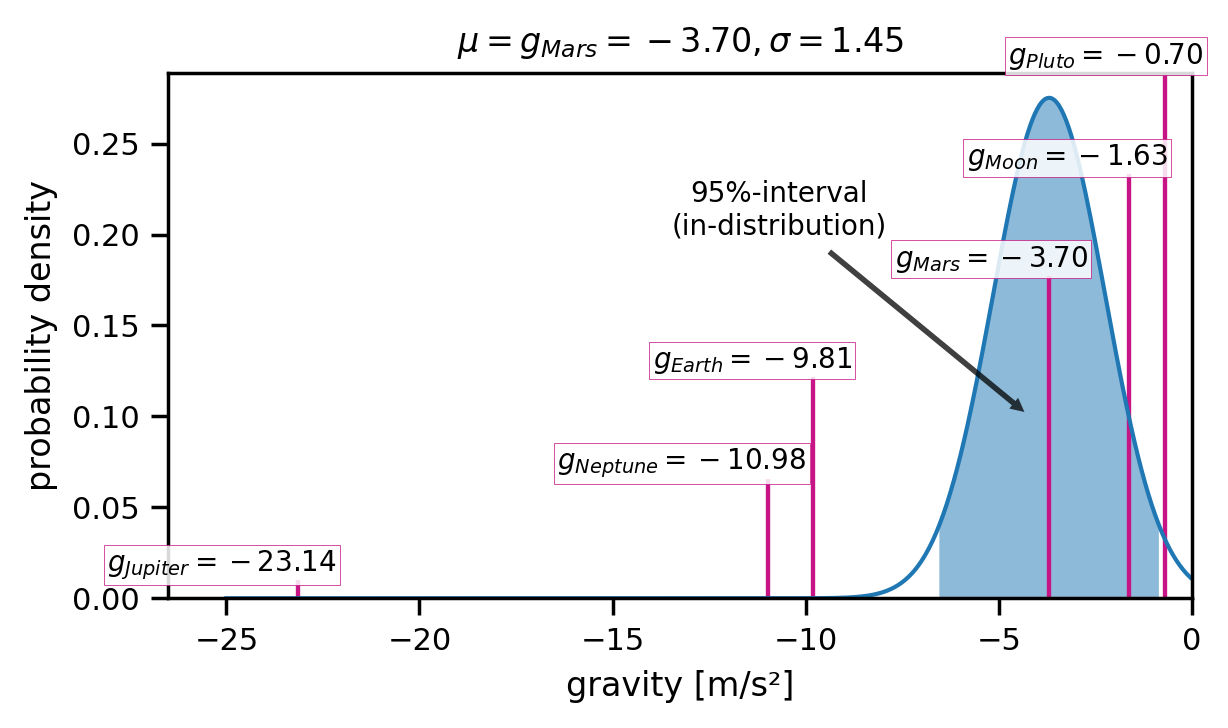}
        \caption{Train distribution and test instances (planets)}
        \label{fig:policytransfer_traindistribution}
 	\end{subfigure}
	\begin{subfigure}[t]{0.49\textwidth}
		\centering
    	\includegraphics[width=1\textwidth]{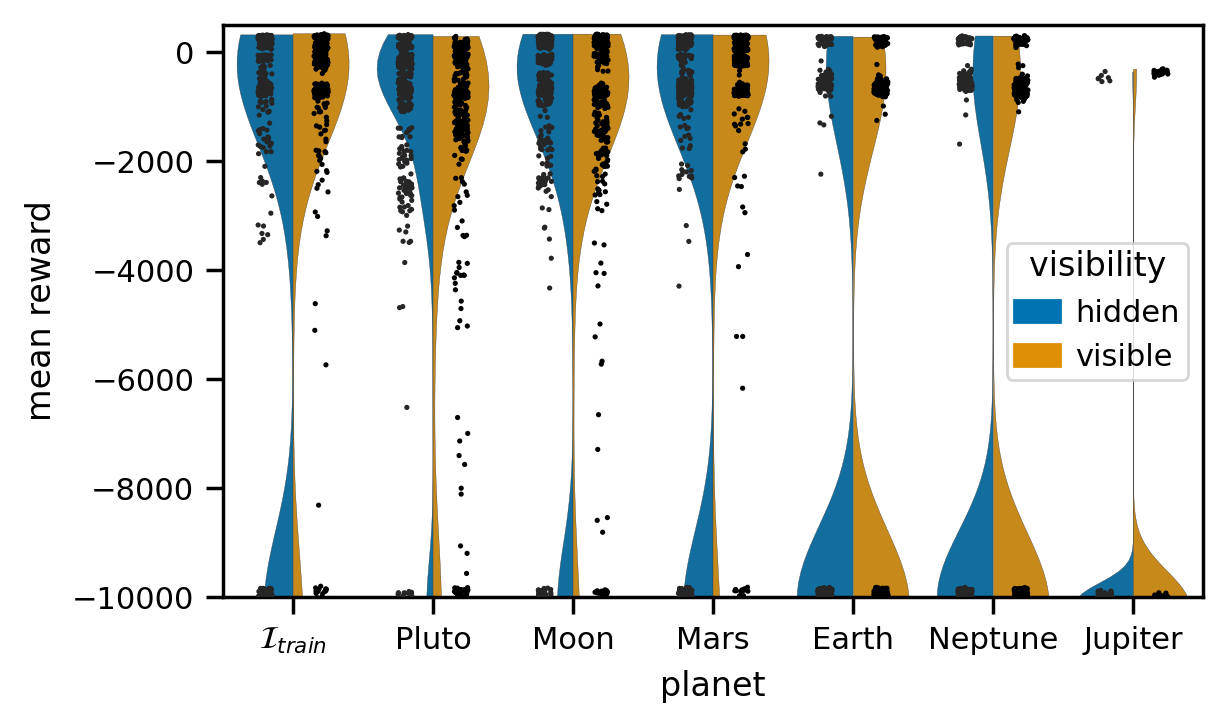}
        \caption{Reward achieved on different instances}\label{fig:results_policytransfer}%
	\end{subfigure}
    \caption{``Landing in Space'' scenario: (\protect\subref{fig:policytransfer_traindistribution}) Mars-centered gravity train distribution with in- and out-of-distribution test instances (planets). (\protect\subref{fig:results_policytransfer}) Resulting rewards for landing on different planets.} 
    \label{fig:policytransfer}
\end{figure}

\paragraph{Landing in Space} In this task, the agent is challenged to land a spacecraft on seen as well as unseen planets.
We model the different planets by only adjusting the gravities by a well-defined normal distribution and train the LunarLander to land on smaller planets.
The train distribution is centered on Mars ($\mu = \SI{3.7}{\metre\per\second^2}$) and the standard deviation ($\sigma = \SI{1.45}{\metre\per\second^2}$) is chosen such that Mars and Moon are considered as in-distribution whilst Pluto, Earth, Neptune and Jupiter are considered as unseen and out-of-distribution,
see Figure~\ref{fig:policytransfer_traindistribution}. 
Here we deem planets as \emph{in-distribution} if their gravity is within the $\SI{95}{\percent}$-interval of the training distribution.
For training, we sample $100$ gravities from this distribution.
We use $5$ random seeds for training and testing and collect $100$ episodes on each planet for both cases where the context is hidden and where the context feature gravity is visible to the agent.
Note that although for each test episode on a planet the gravity is fixed and the same, the LunarLander environment generates different initial starting conditions and landscapes to land on. For this reason the lander might still fail to safely land and crash in some cases.
To capture crashes and to distinguish them from successful but suboptimal landings, we increase the game over penalty from $-100$ to $-10000$ during testing. 

\paragraph{In Distribution Generalization} As to be expected, the test performances for landing on Mars and Moon are most similar to the Mars-centered training distribution. Agents trained with access to the gravity feature receive higher rewards and less crashes on in-distribution planets than their context-oblivious counterparts, as shown at the bottom of Figure~\ref{fig:results_policytransfer} .

\paragraph{Out of Distribution Generalization} A more interesting and understudied question in RL is the extend to which agents are capable of generalizing to out-of-distribution tasks.
With \GenRL{}s possibility to define particular distributions over context features, we can study this question in detail.
The agent fails least often on Pluto since (a) it has a gravity still close to the training distribution and (b) has a low gravity.
The lower the gravity, the longer the timeframe to land is which creates easier landing scenarios.
We can further note that fewer runs on Pluto lead to as high rewards as on in-distribution planets Moon and Mars.
This is likely due to agents wasting fuel by anticipating a harder landing, thus burning more fuel to counteract.
Interestingly, even for more difficult out-of-distribution planets such as Earth and Neptune we can observe positive landing results for both agents trained with and those without access to the context.
However, test performance deteriorates due to more frequent crashing on more high-gravity planets.
While we have seen some capability of trained agents to transfer even to out-of-distribution environments, we do not expect vanilla agents to generalize to highly different environments.

\section{Further Open Challenges Enabled by \GenRL} Although these experiments only give a first impression on how \GenRL can be used to gain novel insights into RL agents, we see many more possibilities for future research involving \GenRL.
We discuss six open challenges and how \GenRL could be used to tackle these.
(I) As \GenRL provides ground truths for all considered context features it is suitable to study novel agents that separate representation learning from policy learning.
(II) \GenRL will be useful in studying RL agents for uncertain dynamics, by easily perturbing context features.
(III) It is particularly suitable for training and evaluation of continual RL methods by continuously adapting context distributions over time.
(IV) The ground truth on contexts can also be used to study explainability and interpretability methods of deep RL.
(V) With the complexity of modern RL methods, they have become very sensitive to their hyperparameters. \GenRL' flexibility and focus on generalization enables research into AutoRL methods that optimize agents for generality.
(VI) Finally, it is an open question for safe RL whether context information could contribute to decide whether policies are applicable to unseen instances of an environment.
Please find a detailed discussion in Appendix~\ref{appendix:challenge}.

\section{Limitations and Societal and Ethical Implications}
\label{sec:limitations}
Although in principle some environments of \GenRL allow to study the impact of context on vision-based agents, our analysis focuses on featurized environments.
Thus, we did not study different ways of directly exposing context information to vision-based agents which would require novel architectures to handle this context.
We see such experiments and design of novel agents as future work that can follow from using \GenRL.

Our experimentation limits itself to static contexts and does not consider learning with dynamic contexts or continual learning.
We leave this for future studies since generalization to fixed contexts already poses a major challenge.
Lastly, we limited our experiments on varying individual context features.
Off-the-shelf agents are not yet designed to be adaptive to contexts.
Varying individual features already posed a challenge to learn with for the considered agents.
With progress in the field we hope that agents will become more flexible and can handle ever more changes in environments.

We foresee no new direct societal and ethical implications other than the known concerns regarding autonomous agents and RL (e.g., in a military context).
However, by trying to lower the barrier of entry for Meta-RL research we hope to i) reduce the required compute for future research, ii) facilitate novel designs of RL agents and iii) reach a more diverse research community.

\section{Conclusion}
We introduced \GenRL, a highly flexible benchmark library for enabling  studies on generalizable RL via task variations and  context features. 
By employing contextual RL, \GenRL extends common RL environments by making the context configurable and potentially visible. 
Besides providing a ready-to-use benchmark library and discussing the role of context in general RL, we ran first experiments to analyse its aspects.
Our main insights are that (i) the more the context is varied, the more difficult learning becomes and (ii) making the agent context-aware can facilitate training and increase generalization.
In addition, \GenRL is suitable to study generalization in detail by being able to carefully set instance and context distributions. We provide empirical evidence that current agents can generalize well on in-distribution test instances but fail to do so on out-of-distribution settings.
In conclusion, we believe that \GenRL will be a valuable benchmark to advance on open challenges like generalizing RL, representation and continual learning, safe RL and AutoRL. 



\section*{Acknowledgements}
Carolin Benjamins, Theresa Eimer and Marius Lindauer acknowledge funding by the German Research Foundation under LI 2801/4-1.
André Biedenkapp and Frank Hutter acknowledge funding by the Robert Bosch GmbH.

\bibliographystyle{apalike}
\bibliography{bib/meta-gym,bib/strings,bib/lib,bib/proc}

\appendix

\section{Benchmark Categories}\label{appendix:cats}
To encourage generalization in RL, we chose a wide variety of common task characteristics as well as well-known environments as the basis of \GenRL.

The physical simulation environments (Brax, box2d and classic control) defining a dynamic body in a static world have similar context features like gravity, geometry of the moving body, position and velocity, mass, friction and joint stiffness.
For brevity, we only detail the context features of CARLFetch and list all other environments' context features in Section~\ref{sec:appendix_env_context_features} of the appendix.

CARLFetch embeds Brax' Fetch~\cite{brax2021github} as a cMDP, see Figure~\ref{fig:fetch_cMDP}.
The goal of Fetch is to move the agent to the target area.
The context features joint stiffness, gravity, friction, (joint) angular damping, actuator strength, torso mass as well as target radius and distance define the context.
The defaults of the context features are copied from the original environment.
Furthermore, appropriate bounds must be set for the specific application. We set the bounds such that the environment's purpose is not violated, e.g., restricting the gravity towards the ground greater than $0$ (otherwise the agent would fly up and it would be impossible to act).

Besides physical simulation environments, \GenRL provides two more specific, challenging environments.
The first is the CARLMarioEnv environment built on top of the TOAD-GAN level generator \cite{awiszus-aiide20,schubert-tg21}. 
It provides a procedurally generated game playing environment (similarly to the ones discussed in Section~\ref{sec:context_matters}) that allows customization of the generation process. 
This environment is therefore especially interesting for exploring  representation learning for the purpose of learning to better generalize.
Secondly, we move closer to real-world application by including the \mbox{CARLRNADesignEnvironment}~\cite{runge-iclr19a}. 
The task here is to design RNA sequences given structural constraints.
As two different datasets of structures and their instances are used in this benchmark, it is ideally suited for testing policy transfer between RNA structures.

\subsection{Pendulum's Dynamic Equations}
\label{sec:pendulum_dyna_eq}
Because we use gym's Pendulum~\cite{gym} for our experiments Q1 and Q2 (see section~\ref{sec:experiments}), we provide the dynamic equations to show the simplicity of the system.
The state consists of the angular position $\theta$ and velocity of the pendulum $\dot{\theta}$.
The discrete equation defining the behavior of the environment is defined as follows:
\begin{align*}\label{eq:pendulum_eq}
\dot{\theta}_{k+1} &= \dot{\theta}_k + (- \frac{3g}{2l} \sin(\theta_k + \pi) + \frac{3}{m \cdot l^2}u_k) \cdot \Delta t \\
\theta_{k+1} &= \theta_k + \dot \theta_{k+1} \cdot \Delta t \,.
\end{align*}
Here, $k$ is the index of the iteration/step, $g$ the gravity, $l$ the length of the pendulum, $u$ the control input and $\Delta t$ the timestep.

\section{Hardware and Hyperparameters}\label{appendix:hardhyper}

\paragraph{Hardware}
All experiments on all benchmarks were conducted on a slurm CPU cluster if not stated otherwise (see Table \ref{tab:cpu_cluster}). The experiments for CARLMarioEnv were replicated on a slurm GPU cluster consisting of 6 nodes with eight RTX 2080 Ti each.

\begin{table}[ht]
    \centering
    \caption{CPU cluster used for training}
    \label{tab:cpu_cluster}
    \begin{tabular}{c|c|c|c}
         Machine no. &  CPU model & cores & RAM \\
         \hline
          1 &  Xeon E5-2670 & 16 & 188 GB \\
          2 & Xeon E5-2680 v3 & 24 & 251 \\
          3-6 & Xeon E5-2690 v2 & 20 & 125 GB \\
          7-10 &  Xeon Gold 5120 & 28 & 187 \\
    \end{tabular}

\end{table}

\paragraph{Hyperparameters and Training Details} We used agents from stable baselines 3 \cite{stable-baselines3} (version 1.1.0) for all of our experiments. For the DQN (used in CARLLunarLanderEnv, Section~\ref{sec:policy_transfer}) and the PPO agent (used in CARLMarioEnv in Section~\ref{sec:additional_results}) we employ the hyperparameters from the stable baselines zoo~\cite{rl-zoo3}, see Table~\ref{tab:agent_hps}. For the DDPG agent (used for CARLPendulumEnv in Section~\ref{sec:context_influence} and~\ref{sec:hidden_vs_visible_context}) we use the defaults with a MLP policy. We train each agent for $10^6$ steps. Every $5000$ steps we evaluate one episode on each train instance and report the mean reward across instances.
All experiments can be reproduced using the scripts we provide with the benchmark library at \mbox{\url{https://www.github.com/automl/CARL}}.
\begin{table}[ht]
    \centering
    \caption{Hyperparameters from stable baselines zoo~\cite{rl-zoo3} for the agents used. Blank fields mean default values from stable baselines agent.}
    \label{tab:agent_hps}
    \small
    \begin{tabular}{l r r}
    \toprule
    Hyperparameter          & DQN                           & PPO  \\
    \midrule
    n\_envs                  & 1                             & 8  \\
    policy                  & MlpPolicy                     & CnnPolicy  \\
    n\_steps                 &                               & 128  \\
    n\_epochs                &                               & 4  \\
    learning\_rate           & 6.3e-4                        & lin\_2.5e-4  \\
    batch\_size              & 128                           & 256  \\
    n\_epochs                & 4                             &  \\
    buffer\_size             & 50000                         &  \\
    learning\_starts         & 0                             &  \\
    gamma                   & 0.99                          &  \\
    target\_update\_interval  & 250                           &  \\
    train\_freq              & 4                             &  \\
    gradient\_steps          & -1                            &  \\
    exploration\_fraction    & 0.12                          &  \\
    exploration\_final\_eps   & 0.1                           &  \\
    clip\_range              &                               & lin\_0.1  \\
    policy\_kwargs           & net\_arch=[256, 256]           &  \\
    vf\_coef                 &                               & 0.5  \\
    ent\_coef                &                               & 0.01  \\
    \bottomrule
    \end{tabular}
\end{table}

\section{Additional Experimental Results}
\label{sec:additional_results}
\begin{figure}[ht!]
    \centering
    \includegraphics[width=0.7\textwidth]{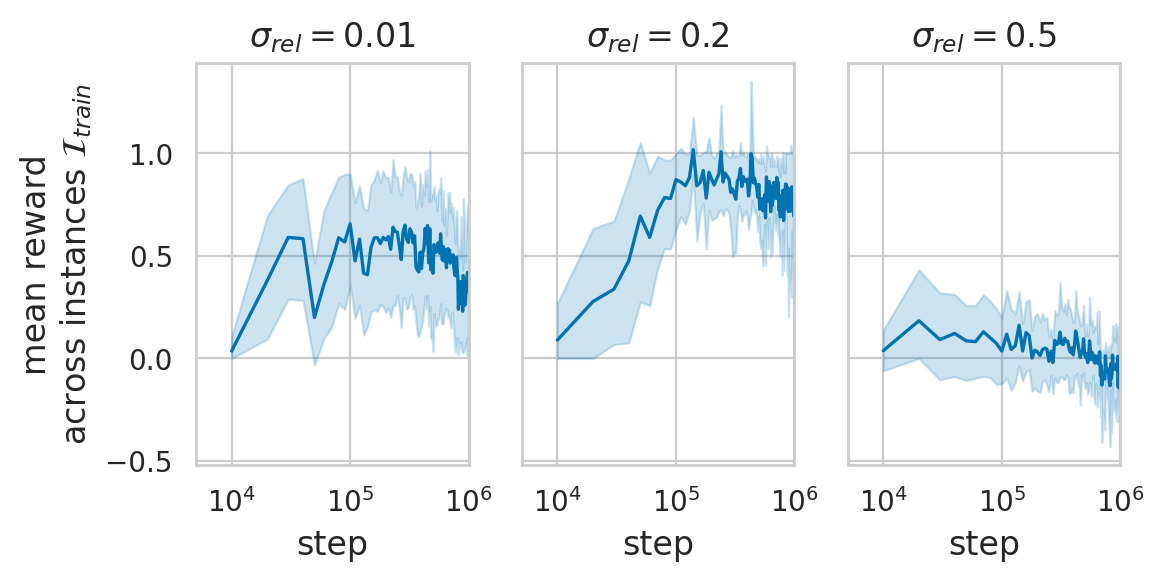}
    \caption{Training performance on CARLMarioEnv where only noise on the input generative vector is changed.}
    \label{fig:results_mario}
\end{figure}

To further illustrate the influence of varying context, we show experimental results for a PPO agent trained on the CARLMarioEnv.
Again, the agent is trained for $5$ different random seeds and with $100$ training instances.
In CARLMarioEnv different instances (Mario levels) are created by using TOAD-GAN~\cite{schubert-tg21}.
By varying the noise input vector for TOAD-GAN we can generate different levels and the greater the noise, the greater the differences to the original level.
Because CARLMarioEnv is pixel-based the context is implicitly coded in the state and we hide the context.
As we can see in Figure~\ref{fig:results_mario} a diverse training distribution ($\sigma_\text{rel}=0.2$) increases the performance and facilitates generalization.
On the other hand if the noise becomes too large ($\sigma_\text{rel}=0.5$) the performance decreases again.
A reason might be that the levels for this noise level are very diverse and thus the current setup, with only implicit context, might not be suitable. 


\section{Open Challenges Enabled by \GenRL}\label{appendix:challenge}
We used \GenRL to demonstrate the usefulness of a benchmark that can provide the ground truth of available context information.
Based on that, we showed that making such information about the environment explicitly available to the agent enables faster training and transfer of agents (see Section~\ref{sec:experiments}).
While this already provides valuable insights to the community that increasingly cares about learning agents capable of generalization (see Sections \ref{sec:intro} \& \ref{sec:related_work}) CARL enables to study further open challenges for general RL.

\subsection{Challenge I: Representation Learning}\label{subsec:repl}
Our experiments using \GenRL demonstrated that an agent that is given access to context information is capable of learning better than an agent that has to learn behaviours given an implicit context via state observations.
This provides evidence that disentangling the representation learning aspect from the policy learning task reduces complexity.
As \GenRL provides a ground truth for representations of environment properties we envision future work on principled studies of novel RL algorithms that, by design, disentangle representation learning and policy learning (see, e.g., \cite{rakelly-icml19,fu-aaai21,zhang-iclr21a} as first works along this line of research).
The ground truth given by the context would allow to measure the quality of learned representations and allows us to relate this to true physical properties of an environment.

Another use-case of \GenRL we envision under the umbrella of representation learning follows the work of environment probing policies~\cite{zhou-iclr19}.
There, exploratory policies are learned that allow to identify which environment type an agent encounters.
This is complementary to the prior approaches as representations are not jointly learned with the behaviour policies as in the previously discussed approaches but rather in a separate offline phase.
Based on \GenRL, huge amounts of meta-data could be collected that will enable the community to make use of classical meta-algorithmic approaches such as algorithm selection~\cite{rice76a} for selecting previously learned policies or learning approaches. 

\subsection{Challenge II: Uncertainty of RL Agents}
With the access to context information \GenRL enables to study the influence of noise on RL agents in a novel way.
While prior environments enabled studies of the behaviour of agents when they could not be certain about their true state in a particular environment, \GenRL further allows to study agents behaviours in scenarios with uncertainty on their current contextual environment, e.g., because of noise on the context features.
In practical deployment of RL, this is reasonable concern since context feature have to be measured somehow by potentially noisy censors.
As this setting affects the overall transition dynamics, \GenRL provides a unique test-bed in which the influence of uncertainty can be studied and how RL agents can deal with such.

\subsection{Challenge III: Continual Learning}
With the flexibility and easy modifiability of \GenRL{}s provided contexts, \GenRL is suitable for studying continual reinforcement learning agents.
In this setting, the distributions provided by \GenRL could be modified, e.g., gradually  shifted, during the training procedure.
For example, \GenRL could be used to evaluate the behaviour of an agent in the Brax environments where one or more joints become stiffer over time.
A learning agent would need to be able to handle this and adapt its gait accordingly.
In particular, one could at some point ``repair'' the agent and reset the joints to their original stiffness.
This would then allow to evaluate whether the agent has ``unlearned'' the original gait.
In the same way, \GenRL allows also to study how agents would react to spontaneous, drastic changes, e.g., broken legs or changes of the environment such as changes of weather conditions. 

\subsection{Challenge IV: Interpretable and Explainable Deep RL}
Trust is a crucial factor, for which interpretability or explainability often is mandatory.
With the provided ground truth through the explicit use of context features, \GenRL could be the base for studying interpretability and explainability of (deep) RL.
By enabling AutoRL studies and different representation learning approaches, \GenRL will contribute to better interpret the training procedures.

\GenRL further allows to study explainability on the level of learned policies.
We propose to study the sensitivity of particular policies to different types of context.
Thus, the value and variability of a context might serve as a proxy to explain the resulting learned behavior.
Such insights might then be used to predict how policies might look like or act (e.g., in terms of frequency of action usage) in novel environments, solely based on the provided context features.

\subsection{Challenge V: AutoRL}\label{subsec:autorl}
AutoRL (e.g., \cite{jaderberg-arxiv17a,runge-iclr19a,parkerholder-neurips20,coreyes-iclr21}) addresses the optimization of the RL learning process.
To this end, hyperparameters, architectures or both of agents are adapted either on the fly~\cite{jaderberg-arxiv17a} or once at the beginning of a run~\cite{runge-iclr19a}.
However, as AutoRL typically requires large compute resources for this procedure, optimization is most often done only on a per-environment basis.
It is reasonable to assume that such hyperparameters might not transfer well to unseen environments, as the learning procedures were not optimized to be robust or to facilitate generalization, but only to improve the reward on a particular instance.

As \GenRL provides easy-to-use contextual extensions of a diverse set of RL problems, it could be used to drive research in this open challenge of AutoRL.
First of all, it enables a large scale-study to understand how static and dynamic configuration approaches complement each other and when one approach is to be preferred over another.
Such a study will most likely also lead to novel default hyperparameter configurations that are more robust and tailored to fast learning and good generalization.
In addition, it will open up the possibility to study whether it is reasonable to use a single hyperparameter configurations or whether a mix of configurations for different instances is required~\cite{xu-aaai10a}.
Furthermore, with the flexibility of defining a broad variety of instance distributions for a large set of provided context features, experiments with \GenRL would allow researchers to study which hyperparameters play a crucial role in learning general agents similar to studies done for supervised machine learning~\cite{rijn-kdd18a} or AI algorithms~\cite{biedenkapp-lion18a}.


\subsection{Challenge VI: High Confidence Generalization}
The explicit context of the \GenRL{} benchmark enables tackling another challenge in the field of safe RL.
High Confidence Generalization algorithms (HCGAs)~\cite{kostas-pmlr21} provide safety guarantees for the generalization of agents in testing environments.
Given a worst-case performance bound, the agent can be tasked to decide whether a policy is applicable in an out-of-distribution context or not.
This setting is especially important for the deployment of RL algorithms in the real world where policy failures can be costly and the context of an environment is often prone to change.
\GenRL{} has the potential to facilitate the development of HCGAs that base their confidence estimates on the context of an environment.

\section{Future Maintenance}
As our benchmark draws from several different RL environments as dependencies, we realize that it will need regular maintenance and updating.
Furthermore, we would like to include more benchmarks and options that are closer to real-world applications.
In part, we of course hope that the community will embrace \GenRL and work with us to extend it in order to match the needs of researchers working in cRL.
We acknowledge, however, that relying on community driven progress only is infeasible.
Therefore we commit to updating the current benchmark version including its dependencies at least twice a year or whenever critical updates in dependencies are released. 
As we plan to continue using GitHub for hosting, versioning as well as providing continued access to previous versions is feasible.
We also aim to fix any issues that are brought to our attention in a reasonable timeframe.
In case community-driven benchmarks are added, we will ensure the continued functionality of the benchmark as a whole (as far as our resources will allow).
As we are researching solution methods in the field of cRL ourselves, we expect to contribute further benchmarks of our own as well.

\section{Statement}
The authors' acknowledge that they bear all responsibility in case of violation of rights, etc., and confirmation of the data license.

\section{Context Features for Each Environment}
\label{sec:appendix_env_context_features}

We list all registered context features with their defaults, bounds and types for each environment family in Table~\ref{tab:cf_defs_classiccontrol} (classic control), Table~\ref{tab:cf_defs_box2d} (box2d), Table~\ref{tab:cf_defs_brax} (brax) and Table~\ref{tab:cf_defs_misc} (RNA and Mario).

\begin{table}[ht!]
    \caption{Context Features: Defaults, Bounds and Types for OpenAI gym's Classic Control environments~\cite{gym}}
    \label{tab:cf_defs_classiccontrol}

    \small{\begin{subtable}{0.4\textwidth}
\centering
\caption[CARLCartPoleEnv]{CARLCartPoleEnv}
\label{tab:context_features_defaults_bounds_CARLCartPoleEnv}
\begin{tabular}{lrll}
\toprule
Context Feature &  Default &       Bounds &  Type \\
\midrule
force\_magnifier &    10.00 &     (1, 100) &   int \\
        gravity &     9.80 &   (0.1, inf) & float \\
       masscart &     1.00 &    (0.1, 10) & float \\
       masspole &     0.10 &    (0.01, 1) & float \\
    pole\_length &     0.50 &    (0.05, 5) & float \\
update\_interval &     0.02 & (0.002, 0.2) & float \\
\bottomrule
\end{tabular}
\end{subtable}
}
    \hspace{0.5in}
    \small{\begin{subtable}{0.4\textwidth}
\centering
\caption[CARLPendulumEnv]{CARLPendulumEnv}
\label{tab:context_features_defaults_bounds_CARLPendulumEnv}
\begin{tabular}{lrll}
\toprule
Context Feature &  Default &       Bounds &  Type \\
\midrule
             dt &     0.05 &     (0, inf) & float \\
              g &    10.00 &     (0, inf) & float \\
              l &     1.00 & (1e-06, inf) & float \\
              m &     1.00 & (1e-06, inf) & float \\
      max\_speed &     8.00 &  (-inf, inf) & float \\
\bottomrule
\end{tabular}
\end{subtable}
}

    \small{\begin{subtable}{0.4\textwidth}
\centering
\caption[CARLMountainCarEnv]{CARLMountainCarEnv}
\label{tab:context_features_defaults_bounds_CARLMountainCarEnv}
\begin{tabular}{lrll}
\toprule
   Context Feature &  Default &      Bounds &  Type \\
\midrule
             force &     0.00 & (-inf, inf) & float \\
     goal\_position &     0.50 & (-inf, inf) & float \\
     goal\_velocity &     0.00 & (-inf, inf) & float \\
           gravity &     0.00 &    (0, inf) & float \\
      max\_position &     0.60 & (-inf, inf) & float \\
         max\_speed &     0.07 &    (0, inf) & float \\
      min\_position &    -1.20 & (-inf, inf) & float \\
    start\_position &    -0.50 & (-1.5, 0.5) & float \\
start\_position\_std &     0.10 &  (0.0, inf) & float \\
    start\_velocity &     0.00 & (-inf, inf) & float \\
start\_velocity\_std &     0.00 &  (0.0, inf) & float \\
\bottomrule
\end{tabular}
\end{subtable}
}
    \hspace{0.5in}
    \small{\begin{subtable}{0.4\textwidth}
\centering
\caption[CARLAcrobotEnv]{CARLAcrobotEnv}
\label{tab:context_features_defaults_bounds_CARLAcrobotEnv}
\begin{tabular}{lrll}
\toprule
Context Feature &  Default &                                   Bounds &  Type \\
\midrule
     link\_com\_1 &     0.50 &                                   (0, 1) & float \\
     link\_com\_2 &     0.50 &                                   (0, 1) & float \\
  link\_length\_1 &     1.00 &                                (0.1, 10) & float \\
  link\_length\_2 &     1.00 &                                (0.1, 10) & float \\
    link\_mass\_1 &     1.00 &                                (0.1, 10) & float \\
    link\_mass\_2 &     1.00 &                                (0.1, 10) & float \\
       link\_moi &     1.00 &                                (0.1, 10) & float \\
 max\_velocity\_1 &    12.57 & (1.26, 125.66) & float \\
 max\_velocity\_2 &    28.27 &   (2.83, 282.74) & float \\
\bottomrule
\end{tabular}
\end{subtable}
}

\end{table}

\begin{table}[ht!]
    \caption{Context Features: Defaults, Bounds and Types for OpenAI gym's Box2d environments~\cite{gym}}
    \label{tab:cf_defs_box2d}
    
    \scriptsize{\begin{subtable}{0.4\textwidth}
\centering
\caption[CARLBipedalWalkerEnv]{CARLBipedalWalkerEnv}
\label{tab:context_features_defaults_bounds_CARLBipedalWalkerEnv}
\begin{tabular}{lrll}
\toprule
 Context Feature &  Default &       Bounds &  Type \\
\midrule
             FPS &    50.00 &     (1, 500) & float \\
        FRICTION &     2.50 &      (0, 10) & float \\
       GRAVITY\_X &     0.00 &    (-20, 20) & float \\
       GRAVITY\_Y &   -10.00 & (-20, -0.01) & float \\
  INITIAL\_RANDOM &     5.00 &      (0, 50) & float \\
        LEG\_DOWN &    -0.27 &  (-2, -0.25) & float \\
           LEG\_H &     1.13 &    (0.25, 2) & float \\
           LEG\_W &     0.27 &  (0.25, 0.5) & float \\
     LIDAR\_RANGE &     5.33 &    (0.5, 20) & float \\
   MOTORS\_TORQUE &    80.00 &     (0, 200) & float \\
           SCALE &    30.00 &     (1, 100) & float \\
       SPEED\_HIP &     4.00 &  (1e-06, 15) & float \\
      SPEED\_KNEE &     6.00 &  (1e-06, 15) & float \\
   TERRAIN\_GRASS &    10.00 &      (5, 15) &   int \\
  TERRAIN\_HEIGHT &     5.00 &      (3, 10) & float \\
  TERRAIN\_LENGTH &   200.00 &   (100, 500) &   int \\
TERRAIN\_STARTPAD &    20.00 &     (10, 30) &   int \\
    TERRAIN\_STEP &     0.47 &    (0.25, 1) & float \\
      VIEWPORT\_H &   400.00 &   (200, 800) &   int \\
      VIEWPORT\_W &   600.00 &  (400, 1000) &   int \\
\bottomrule
\end{tabular}
\end{subtable}
}
    \hspace{0.5in}
    \scriptsize{\begin{subtable}{0.4\textwidth}
\centering
\caption[CARLLunarLanderEnv]{CARLLunarLanderEnv}
\label{tab:context_features_defaults_bounds_CARLLunarLanderEnv}
\begin{tabular}{lrll}
\toprule
   Context Feature &  Default &       Bounds &  Type \\
\midrule
               FPS &    50.00 &     (1, 500) & float \\
         GRAVITY\_X &     0.00 &    (-20, 20) & float \\
         GRAVITY\_Y &   -10.00 & (-20, -0.01) & float \\
    INITIAL\_RANDOM &  1000.00 &    (0, 2000) & float \\
          LEG\_AWAY &    20.00 &      (0, 50) & float \\
          LEG\_DOWN &    18.00 &      (0, 50) & float \\
             LEG\_H &     8.00 &      (1, 20) & float \\
 LEG\_SPRING\_TORQUE &    40.00 &     (0, 100) & float \\
             LEG\_W &     2.00 &      (1, 10) & float \\
 MAIN\_ENGINE\_POWER &    13.00 &      (0, 50) & float \\
             SCALE &    30.00 &     (1, 100) & float \\
  SIDE\_ENGINE\_AWAY &    12.00 &      (1, 20) & float \\
SIDE\_ENGINE\_HEIGHT &    14.00 &      (1, 20) & float \\
 SIDE\_ENGINE\_POWER &     0.60 &      (0, 50) & float \\
        VIEWPORT\_H &   400.00 &   (200, 800) &   int \\
        VIEWPORT\_W &   600.00 &  (400, 1000) &   int \\
\bottomrule
\end{tabular}
\end{subtable}
}
    
    \small{\begin{subtable}{0.4\textwidth}
\centering
\caption[CARLVehicleRacingEnv]{CARLVehicleRacingEnv}
\label{tab:context_features_defaults_bounds_CARLVehicleRacingEnv}
\begin{tabular}{lrll}
\toprule
Context Feature &  Default & Bounds &        Type \\
\midrule
        VEHICLE &        0 &      - & categorical, $n=29$ \\
\bottomrule
\end{tabular}
\end{subtable}
}
\end{table}

\begin{table}[ht!]
    \caption{Context Features: Defaults, Bounds and Types for Google Brax environments~\cite{brax2021github}}
    \label{tab:cf_defs_brax}
    
    \scriptsize{\begin{subtable}{0.4\textwidth}
\centering
\caption[CARLAnt]{CARLAnt}
\label{tab:context_features_defaults_bounds_CARLAnt}
\begin{tabular}{lrll}
\toprule
      Context Feature &  Default &       Bounds &  Type \\
\midrule
    actuator\_strength &   300.00 &     (1, inf) & float \\
      angular\_damping &    -0.05 &  (-inf, inf) & float \\
             friction &     0.60 &  (-inf, inf) & float \\
              gravity &    -9.80 & (-inf, -0.1) & float \\
joint\_angular\_damping &    35.00 &     (0, inf) & float \\
      joint\_stiffness &  5000.00 &     (1, inf) & float \\
           torso\_mass &    10.00 &   (0.1, inf) & float \\
\bottomrule
\end{tabular}
\end{subtable}
}
    \hspace{0.5in}
    \scriptsize{\begin{subtable}{0.4\textwidth}
\centering
\caption[CARLFetch]{CARLFetch}
\label{tab:context_features_defaults_bounds_CARLFetch}
\begin{tabular}{lrll}
\toprule
      Context Feature &  Default &       Bounds &  Type \\
\midrule
    actuator\_strength &   300.00 &     (1, inf) & float \\
      angular\_damping &    -0.05 &  (-inf, inf) & float \\
             friction &     0.60 &  (-inf, inf) & float \\
              gravity &    -9.80 & (-inf, -0.1) & float \\
joint\_angular\_damping &    35.00 &     (0, inf) & float \\
      joint\_stiffness &  5000.00 &     (1, inf) & float \\
      target\_distance &    15.00 &   (0.1, inf) & float \\
        target\_radius &     2.00 &   (0.1, inf) & float \\
           torso\_mass &     1.00 &   (0.1, inf) & float \\
\bottomrule
\end{tabular}
\end{subtable}
}
    
    \scriptsize{\begin{subtable}{0.4\textwidth}
\centering
\caption[CARLGrasp]{CARLGrasp}
\label{tab:context_features_defaults_bounds_CARLGrasp}
\begin{tabular}{lrll}
\toprule
      Context Feature &  Default &       Bounds &  Type \\
\midrule
    actuator\_strength &   300.00 &     (1, inf) & float \\
      angular\_damping &    -0.05 &  (-inf, inf) & float \\
             friction &     0.60 &  (-inf, inf) & float \\
              gravity &    -9.80 & (-inf, -0.1) & float \\
joint\_angular\_damping &    50.00 &     (0, inf) & float \\
      joint\_stiffness &  5000.00 &     (1, inf) & float \\
      target\_distance &    10.00 &   (0.1, inf) & float \\
        target\_height &     8.00 &   (0.1, inf) & float \\
        target\_radius &     1.10 &   (0.1, inf) & float \\
\bottomrule
\end{tabular}
\end{subtable}
}
    \hspace{0.5in}
    \scriptsize{\begin{subtable}{0.4\textwidth}
\centering
\caption[CARLHumanoid]{CARLHumanoid}
\label{tab:context_features_defaults_bounds_CARLHumanoid}
\begin{tabular}{lrll}
\toprule
      Context Feature &  Default &       Bounds &  Type \\
\midrule
      angular\_damping &    -0.05 &  (-inf, inf) & float \\
             friction &     0.60 &  (-inf, inf) & float \\
              gravity &    -9.80 & (-inf, -0.1) & float \\
joint\_angular\_damping &    20.00 &     (0, inf) & float \\
           torso\_mass &     8.91 &   (0.1, inf) & float \\
\bottomrule
\end{tabular}
\end{subtable}
}
    
\end{table}

\begin{table}[ht!]
    \caption{Context Features: Defaults, Bounds and Types for RNA Design~\cite{runge-iclr19a} and Mario Environment~\cite{awiszus-aiide20,schubert-tg21}}
    \label{tab:cf_defs_misc}
    \small{\begin{subtable}{0.4\textwidth}
\centering
\caption[CARLRnaDesignEnv]{CARLRnaDesignEnv}
\label{tab:context_features_defaults_bounds_CARLRnaDesignEnv}
\begin{tabular}{lrll}
\toprule
Context Feature &  Default & Bounds &        Type \\
\midrule
        mutation\_threshold     &        5 &      (0.1, inf) & float \\
        reward\_exponent        &        1 &      (0.1, inf) & float \\
        state\_radius           &        5 &      (1, inf) & float \\
        dataset                 & eterna   &    -           & categorical, $n=3$ \\
        target\_structure\_ids & f(dataset)      &   (0, inf) & list of int \\
\bottomrule
\end{tabular}
\end{subtable}

}
    \small{\begin{subtable}{0.4\textwidth}
\centering
\caption[CARLMarioEnv]{CARLMarioEnv}
\label{tab:context_features_defaults_bounds_CARLMarioEnv}
\begin{tabular}{lrll}
\toprule
Context Feature &  Default & Bounds &        Type \\
\midrule
        level\_index &        0 &      - & categorical, $n=15$ \\
        noise       &  f(level\_index, width, height) & (-1, 1) & float \\
        mario\_state & 0 & - & categorical, $n=3$ \\
        mario\_inertia &  0.89 & (0.5, 1.5) & float \\
\bottomrule
\end{tabular}
\end{subtable}

}

\end{table}

\nocite{rl-zoo3}
\nocite{stable-baselines3}
\nocite{rakelly-icml19,fu-aaai21,zhou-iclr19,rice76a}
\nocite{jaderberg-arxiv17a,parkerholder-neurips20,coreyes-iclr21}
\nocite{xu-aaai10a,rijn-kdd18a,biedenkapp-lion18a,kostas-pmlr21}

\end{document}